# Functional Centering

**Michael Strube & Udo Hahn**

Freiburg University
Computational Linguistics Lab
Europaplatz 1, D-79085 Freiburg, Germany
{strube,hahn}@coling.uni-freiburg.de

## Abstract

Based on empirical evidence from a free word order language (German) we propose a fundamental revision of the principles guiding the ordering of discourse entities in the forward-looking centers within the centering model. We claim that grammatical role criteria should be replaced by indicators of the functional information structure of the utterances, i.e., the distinction between context-bound and unbound discourse elements. This claim is backed up by an empirical evaluation of functional centering.

## 1 Introduction

The centering model has evolved as a methodology for the description and explanation of the local coherence of discourse (Grosz et al., 1983; 1995), with focus on pronominal and nominal anaphora. Though several cross-linguistic studies have been carried out (cf. the enumeration in Grosz et al. (1995)), an almost canonical scheme for the ordering on the forward-looking centers has emerged, one that reflects well-known regularities of fixed word order languages such as English. With the exception of Walker et al. (1990; 1994) for Japanese, Turan (1995) for Turkish, Rambow (1993) for German and Cote (1996) for English, only *grammatical roles* are considered and the (partial) ordering in Table 1[1] is taken for granted.

> subject > dir-object > indir-object
> > complement(s) > adjunct(s)

Table 1: Grammatical Role Based Ranking on the $C_f$

[1]Table 1 contains the most explicit ordering of grammatical roles we are aware of and has been taken from Brennan et al. (1987). Often, the distinction between complements and adjuncts is collapsed into the category "others" (c.f., e.g., Grosz et al. (1995)).

Our work on the resolution of anaphora (Strube & Hahn, 1995; Hahn & Strube, 1996) and textual ellipsis (Hahn et al., 1996), however, is based on German, a free word order language, in which grammatical role information is far less predictive for the organization of centers. Rather, for establishing proper referential relations, the *functional information structure* of the utterances becomes crucial (different perspectives on functional analysis are brought forward in Daneš (1974b) and Dahl (1974)). We share the notion of functional information structure as developed by Daneš (1974a). He distinguishes between two crucial dichotomies, viz. *given information* vs. *new information* (constituting the *information structure* of utterances) on the one hand, and *theme* vs. *rheme* on the other (constituting the *thematic structure* of utterances; cf. Halliday & Hasan (1976, pp.325-6)). Daneš refers to a definition given by Halliday (1967) to avoid the confusion likely to arise in the use of these terms: "[...] while *given* means *what you were talking about* (or *what I was talking about before*), *theme* means *what I am talking about (now)* [...]" Halliday (1967, p.212). Daneš concludes that the distinction between *given information* and *theme* is justified, while the distinction between *new information* and *rheme* is not. Thus, we arrive at a trichotomy between *given information*, *theme* and *rheme* (the latter being equivalent to *new information*). We here subscribe to these considerations, too, and will return in Section 3 to these notions in order to rephrase them more explicitly by using the terminology of the centering model.

In this paper, we intend to make two contributions to the centering approach. The first one, the introduction of functional notions of information structure in the centering model, is methodological in nature. The second one concerns an empirical issue in that we demonstrate how a functional model of centering can successfully be applied to the analysis of several forms of anaphoric text phenomena.

At the methodological level, we develop arguments that (at least for free word order languages) grammatical role indicators should be replaced by functional role patterns to more adequately account for the ordering of discourse entities in center lists. In Section 3 we elaborate on the particular information structure criteria underlying a function-based center ordering. We also make a second, even more general methodological claim for which we have gathered some preliminary, though still not conclusive evidence. Based on a re-evaluation of empirical arguments discussed in the literature on centering, we stipulate that exchanging grammatical by functional criteria is also a reasonable strategy for fixed word order languages. Grammatical role constraints can indeed be rephrased by functional ones, which is simply due to the fact that grammatical roles and the information structure patterns, as we define them, coincide in these kinds of languages. Hence, the proposal we make seems more general than the ones currently under discussion in that, given a functional framework, fixed and free word order languages can be accounted for by the same ordering principles. As a consequence, we argue against Walker et al.'s (1994, p.227) stipulation, which assumes that the $C_f$ ranking is the only parameter of the centering theory which is language-dependent. Instead, we claim that *functional centering* constraints for the $C_f$ ranking are possibly universal.

The second major contribution of this paper is related to the unified treatment of specific text phenomena. It consists of an equally balanced treatment of intersentential (pro)nominal anaphora and textual ellipsis (also called functional or partial anaphora). The latter phenomenon (cf. the examples given in the next section), in particular, is usually only sketchily dealt with in the centering literature, e.g., by asserting that the entity in question "is realized but not directly realized" (Grosz et al., 1995, p.217). Furthermore, the distinction between those two kinds of realization is generally delegated to the underlying semantic theory. We will develop arguments how to locate elliptical discourse entities and resolve textual ellipsis properly at the center level. The ordering constraints we supply account for all of the above mentioned types of anaphora in a precise way, including (pro)nominal anaphora (Strube & Hahn, 1995; Hahn & Strube, 1996). This claim will be validated by a substantial body of empirical data (cf. Section 4).

## 2 Types of Anaphora Considered

Text phenomena, e.g., textual forms of ellipsis and anaphora, are a challenging issue for the design of parsers for text understanding systems, since imperfect recognition facilities either result in referentially incoherent or invalid text knowledge representations. At the conceptual level, textual ellipsis relates a quasi-anaphoric expression to its extrasentential antecedent by conceptual attributes (or roles) associated with that antecedent (see, e.g., the relation between *"Akkus" (accumulator)* and *"316LT"*, a particular notebook, in (1b) and (1a)). Thus, it complements the phenomenon of nominal anaphora, where an anaphoric expression is related to its antecedent in terms of conceptual generalization (as, e.g., *"Rechner" (computer)* in (1c) refers to *"316LT"* in (1a) mediated by the textual ellipsis in (1b)). The resolution of text-level nominal (and pronominal) anaphora contributes to the construction of referentially valid text knowledge bases, while the resolution of textual ellipsis yields referentially coherent text knowledge bases.

(1) a. Ein Reserve-Batteriepack versorgt den *316LT* ca. 2 Minuten mit Strom.
   (A reserve battery pack – supplies – the *316LT* – for approximately 2 minutes – with power.)

   b. Der Status des *Akkus* wird dem Anwender angezeigt.
   (The status of the *accumulator* – is – to the user – indicated.)

   c. Ca. 30 Minuten vor der *Entleerung* beginnt der *Rechner* 5 Sekunden zu beepen.
   (Approximately 30 minutes – before the *discharge* – starts – the *computer* – for 5 seconds – to beep.)

   d. 5 Minuten bevor *er* sich ausschaltet, fängt die Low-Battery-LED an zu blinken.
   (5 minutes – before – *it* – itself – turns off – begins – the low-battery-LED – to flash.)

In the case of textual ellipsis, the missing conceptual link between two discourse elements occurring in adjacent utterances must be inferred in order to establish the local coherence of the discourse (for an early statement of that idea, cf. Clark (1975)). In the surface form of utterance (1b) the information is missing that *"Akkus" (accumulator)* links up with *"316LT"*. This relation can only be made explicit if conceptual knowledge about the domain, *viz.* the relation *part-of* between the concepts ACCUMULATOR and 316LT, is available (see Hahn et al. (1996) for a more detailed treatment of text ellipsis resolution).

## 3 Principles of Functional Centering

Within the framework of the centering model (Grosz et al., 1995), we distinguish each utterance's backward-looking center $(C_b(U_n))$ and its forward-looking centers $(C_f(U_n))$. The ranking imposed on the elements of the $C_f$ reflects the assumption that the most highly ranked element of $C_f(U_n)$ – the preferred center $C_p(U_n)$ – is the most preferred antecedent of an anaphoric or elliptical expression in $U_{n+1}$, while the remaining elements are partially ordered according to decreasing preference for establishing referential links. Hence, the most important single construct of the centering model is the ordering of the list of forward-looking centers (Walker et al., 1994).

The main difference between Grosz et al.'s work and our proposal concerns the criteria for ranking the forward-looking centers. While Grosz et al. assume that *grammatical roles* are the major determinant for the ranking on the $C_f$, we claim that for languages with relatively free word order (such as German), it is the *functional information structure (IS)* of the utterance in terms of the context-boundedness or unboundedness of discourse elements. The centering data structures and the notion of context-boundedness can be used to redefine Daneš' (1974a) trichotomy between *given information*, *theme* and *new information* (*rheme*). The $C_b(U_n)$, the most highly ranked element of $C_f(U_{n-1})$ realized in $U_n$, corresponds to the element which represents the *given* information. The *theme* of $U_n$ is represented by the preferred center $C_p(U_n)$, the most highly ranked element of $C_f(U_n)$. The *theme/rheme hierarchy* of $U_n$ is represented by $C_f(U_n)$ which – in our approach – is partly determined by the $C_f(U_{n-1})$: the rhematic elements of $U_n$ are the ones not contained in $C_f(U_{n-1})$ (unbound discourse elements); they express the *new information* in $U_n$. The ones contained in $C_f(U_{n-1})$ and $C_f(U_n)$ (bound discourse elements) are thematic, with the theme/rheme hierarchy corresponding to the ranking in the $C_f$s. The distinction between context-bound and unbound elements is important for the ranking on the $C_f$, since bound elements are generally ranked higher than any other non-anaphoric elements (cf. also Hajičová et al. (1992)).

An alternative definition of *theme* and *rheme* in the context of the centering approach is proposed by Rambow (1993). In his approach the *theme* corresponds to the $C_b$ and the *theme/rheme hierarchy* can be derived from those elements of $C_f(U_{n-1})$ that are realized in $U_n$. Rambow does not distinguish, however, between the *information structure* and the *thematic structure* of utterances, which leads to problems when a change of the criteria for recognizing the thematic structure is envisaged. Our approach is flexible enough to accomodate other conceptions of *theme/rheme* as defined, e.g., by Hajičová et al. (1995), since this change affects only the thematic but not the information structure of utterances.

| bound element(s) $>_{IS_{base}}$ unbound element(s) |
|---|
| anaphora $>_{IS_{bound}}$ (possessive pronoun *xor* elliptical antecedent) $>_{IS_{bound}}$ (elliptical expression *xor* head of anaphoric expression) |
| nom head$_1$ $>_{prec}$ nom head$_2$ $>_{prec}$ ... $>_{prec}$ nom head$_n$ |

Table 2: Functional Ranking Constraints on the $C_f$

The rules holding for the ranking on the $C_f$, derived from a German language corpus, are summarized in Table 2. They are organized into three layers[2]. At the top level, $>_{IS_{base}}$ denotes the basic relation for the overall ranking of information structure *(IS)* patterns. Accordingly, any *context-bound* expression in the utterance $U_{n-1}$ is given the highest preference as a potential antecedent of an anaphoric or elliptical expression in $U_n$ while any unbound expression is ranked next to context-bound expressions.

The second relation depicted in Table 2, $>_{IS_{bound}}$, denotes preference relations dealing exclusively with multiple occurrences of (resolved) anaphora, i.e., *bound elements*, in the preceding utterance. $>_{IS_{bound}}$ distinguishes among *different forms* of *context-bound elements* (viz., anaphora, possessive pronouns and textual ellipses) and their associated preference order. The final element of $>_{IS_{bound}}$ is either the elliptical expression or the head of an anaphoric expression which is used as a possessive determiner, a Saxon genitive, a prepositional or a genitival attribute (cf. the ellipsis in (2c): *"die Ladezeit" (the charge time)* vs. *"seine Ladezeit" (its charge time)* or *"die Ladezeit des Akkus" (the accumulator's charge time)*).

For illustration purposes, consider text fragment (1) and the corresponding $C_b/C_f$ data in Table 3[3]: In (1d) the pronoun *"er" (it)* might be resolved to *"Akku" (accumulator)* or *"Rechner" (computer)*, since both fulfill the agreement condition for pronoun resolution. Now, *"der Rechner" (computer)* figures as a nominal anaphor, already resolved to DELL-316LT, while *"Akku" (accumulator)* is only the antecedent of the elliptical expression *"der Entleerung" (discharge)*. Therefore, the preferred antecedent of *"er" (it)* is determined as *Rechner (computer)*.

The bottom level of Table 2 specifies $>_{prec}$ which covers the preference order for multiple occurrences of the *same type* of any information structure pattern, e.g., the occurrence of two anaphora or two unbound elements (all heads in an utterance are ordered by linear precedence relative to their text position). In sentence (2b), two nominal anaphors occur, *"Akku" (accumulator)* and *"Rechner" (computer)*. The textual ellipsis *"Ladezeit" (charge time)* in (2c) has to be resolved to the most preferred element of the $C_f$ of (2b), *viz.* the entity denoted by *"Akku" (accumulator)* (cf. Table 4). Note that *"Rechner" (computer)* is the subject of the sentence, though it is not the preferred antecedent, since *"Akku" (accumulator)* precedes *"Rechner" (computer)* and is anaphoric as well.

---

[2]Disregarding coordinations, the ordering we propose induces a *strict* ordering on the entities in a center list.

[3]*Minuten (minutes)* is excluded from the $C_f$ for reasons concerning the processing of complex sentences (cf. Strube (1996)).

|      |       |                                                                                                                                                                 |          |
|------|-------|-----------------------------------------------------------------------------------------------------------------------------------------------------------------|----------|
| (1a) | **Cb:** | DELL-316LT: 316LT                                                                                                                                            | CONTINUE |
|      | **Cf:** | [DELL-316LT: 316LT, RESERVE-BATTERY-PACK: Reserve-Batteriepack, TIME-UNIT-PAIR: 2 Minuten, POWER: Strom]                                                      |          |
| (1b) | **Cb:** | DELL-316LT: —                                                                                                                                                | CONTINUE |
|      | **Cf:** | [DELL-316LT: —, ACCU: Akku, STATUS: Status, USER: Anwender]                                                                                                  |          |
| (1c) | **Cb:** | DELL-316LT: Rechner                                                                                                                                          | CONTINUE |
|      | **Cf:** | [DELL-316LT: Rechner, ACCU: —, DISCHARGE: Entleerung, TIME-UNIT-PAIR: 30 Minuten, TIME-UNIT-PAIR: 5 Sekunden]                                                 |          |
| (1d) | **Cb:** | DELL-316LT: er                                                                                                                                               | CONTINUE |
|      | **Cf:** | [DELL-316LT: er, LOW-BATTERY-LED: Low-Battery-LED                                                                                                             |          |

Table 3: Centering Data for Text Fragment (1)

|      |       |                                                                                                                      |              |
|------|-------|----------------------------------------------------------------------------------------------------------------------|--------------|
| (2a) | **Cb:** | DELL-316LT: 316LT                                                                                                 | CONTINUE     |
|      | **Cf:** | [DELL-316LT: 316LT, NIMH-ACCU: NiMH-Akku]                                                                         |              |
| (2b) | **Cb:** | DELL-316LT: Rechner                                                                                               | RETAIN       |
|      | **Cf:** | [NIMH-ACCU: Akku, DELL-316LT: Rechner, TIME-UNIT-PAIR: 4 Stunden, POWER: Strom]                                   |              |
| (2c) | **Cb:** | NIMH-ACCU: —                                                                                                      | SMOOTH-SHIFT |
|      | **Cf:** | [NIMH-ACCU: —, CHARGE-TIME: Ladezeit, TIME-UNIT-PAIR: 1,5 Stunden]                                                |              |

Table 4: Centering Data for Text Fragment (2)

(2)  a. Der *316LT* wird mit einem *NiMH-Akku* bestückt.
(The *316LT* is – with a *NiMH-accumulator* – equipped.)

b. Durch diesen neuartigen *Akku* wird der *Rechner* für ca. 4 Stunden mit Strom versorgt.
(Because of this new type of *accumulator* – is the *computer* – for approximately 4 hours – with power – provided.)

c. Darüberhinaus ist die *Ladezeit* mit 1,5 Stunden sehr kurz.
(Also – is – the *charge time* of 1.5 hours – quite short.)

Given these basic relations, we may formulate the composite relation $>_{IS}$ (Table 5). It states the conditions for the comprehensive ordering of items on $C_f$ (*x* and *y* denote lexical heads).

---
$>_{IS} := \{ (x, y) \mid$
 *if* x and y both represent the same type of IS pattern
  *then* the relation $>_{prec}$ applies to x and y
 *else if* x and y both represent different forms
       of bound elements
   *then* the relation $>_{IS_{bound}}$ applies to x and y
   *else* the relation $>_{IS_{base}}$ applies to x and y $\}$

---

Table 5: Information Structure Relation

## 4 Evaluation

In this section, we first describe the empirical and methodological framework in which our evaluation experiments were embedded, and then turn to a discussion of evaluation results and the conclusions we draw from the data.

### 4.1 Evaluation Framework

The test set for our evaluation experiment consisted of three different text sorts: 15 product reviews from the information technology *(IT)* domain (one of the two main corpora at our lab), one article from the German news magazine *Der Spiegel*, and the first two chapters of a short story by the German writer *Heiner Müller*[4]. The evaluation was carried out manually in order to circumvent error chaining[5]. Table 6 summarizes the total numbers of anaphors, textual ellipses, utterances and words in the test set.

|         | anaphors | ellipses | utterances | words |
|---------|----------|----------|------------|-------|
| IT      | 308      | 294      | 451        | 5542  |
| Spiegel | 102      | 25       | 82         | 1468  |
| Müller  | 153      | 20       | 87         | 867   |
| Σ       | 563      | 339      | 620        | 7877  |

Table 6: Test Set

Given this test set, we compared three major approaches to centering, *viz.* the original model whose ordering principles are based on grammatical role indicators only (the so-called *canonical* model) as characterized by Table 1, an "intermediate" model which can be considered a *naive* approach to free word order languages, and, of course, the *functional* model based on information structure constraints as stated in Table 2. For reasons discussed below, augmented versions of the naive and the canonical approaches will also be considered. They are characterized by the additional constraint that elliptical antecedents are ranked higher than elliptical expressions (short: "ante > express").

---
[4]Liebesgeschichte. In Heiner Müller. *Geschichten aus der Produktion 2*. Berlin: Rotbuch Verlag, pp. 57-63.

[5]A performance evaluation of the current anaphora and ellipsis resolution capacities of our system is reported in Hahn et al. (1996).

For the evaluation of a centering algorithm on naturally occurring text it is necessary to specify how to deal with complex sentences. In particular, methods for the interaction between intra- and intersentential anaphora resolution have to be defined, since the centering model is concerned only with the latter case (see Suri & McCoy (1994)). We use an approach as described by Strube (1996) for the evaluation.

Since most of the anaphors in these texts are nominal anaphors, the resolution of which is much more restricted than that of pronominal anaphors, the rate of success for the whole anaphora resolution process is not significant enough for a proper evaluation of the functional constraints. The reason for this lies in the fact that nominal anaphors are far more constrained by conceptual criteria than pronominal anaphors. So the chance to properly resolve a nominal anaphor, even at lower ranked positions in the center lists, is greater than for pronominal anaphors. While we shift our evaluation criteria away from simple anaphora resolution success data to structural conditions based on the proper ordering of center lists (in particular, we focus on the most highly ranked item of the forward-looking centers) these criteria compensate for the high proportion of nominal anaphora that occur in our test sets. The types of centering transitions we make use of (cf. Table 7) are taken from Walker et al. (1994).

|  | $C_b(U_n) = C_b(U_{n-1})$ OR $C_b(U_{n-1})$ undef. | $C_b(U_n) \neq C_b(U_{n-1})$ |
|---|---|---|
| $C_b(U_n) = C_p(U_n)$ | CONTINUE | SMOOTH-SHIFT |
| $C_b(U_n) \neq C_p(U_n)$ | RETAIN | ROUGH-SHIFT |

Table 7: Transition Types

### 4.2 Evaluation Results

In Table 8 we give the numbers of centering transitions between the utterances in the three test sets. The first column contains those which are generated by the *naive* approach (such a proposal was made by Gordon et al. (1993) as well as by Rambow (1993) who, nevertheless, restricts it to the German middlefield only). We simply ranked the elements of $C_f$ according to their text position. While it is usually assumed that the elliptical expression ranks above its antecedent (Grosz et al., 1995, p.217), we assume the contrary. The second column contains the results of this modification with respect to the *naive* approach. In the third column of Table 8 we give the numbers of transitions which are generated by the *canonical* constraints as stated by Grosz et al. (1995, p.214, 217). The fourth column supplies the results of the same modification as was used for the naive approach, *viz.* elliptical antecedents are ranked higher than elliptical expressions. The fifth column shows the results which are generated by the *functional* constraints from Table 2.

First, we examine the error data for anaphora resolution for the five cases. All approaches have 99 errors in common. These are due to underspecifications at different levels, e.g., the failure to account for prepositional anaphors (16), plural anaphors (8), anaphors which refer to a member of a set (14), sentence anaphors (21), and anaphors which refer to the global focus (12). Only 6 errors of the functional approach are directly caused by an inappropriate ordering of the $C_f$, while the naive approach leads to 10 errors and the canonical to 7. When the antecedent of an elliptical expression is ranked above the elliptical expression itself the error rate of these two augmented approaches increases to 12 and 9, respectively.

We now turn to the distribution of transition types for the different approaches. The centering model assumes a preference order among these transitions, e.g., CONTINUE ranks above RETAIN and RETAIN ranks above SHIFT. This preference order reflects the presumed inference load put on the hearer or speaker to *coherently* decode or encode a discourse. Since the functional approach generates a larger amount of CONTINUE transitions, we interpret this as a first rough indication that this approach provides for more efficient processing than its competitors.

But this reasoning is not entirely conclusive. Counting single occurrences of transition types, in general, does not reveal the entire validity of the center lists. Instead, considering adjacent transition pairs gives a more reliable picture, since depending on the text sort considered (e.g., technical *vs.* news magazine *vs.* literary texts) certain sequences of transition types may be entirely plausible, though they include transitions which, when viewed in isolation, seem to imply considerable inferencing load (cf. Table 8). For instance, a CONTINUE transition which follows a CONTINUE transition is a sequence which requires the lowest processing costs. But a CONTINUE transition which follows a RETAIN transition implies higher processing costs than a SMOOTH-SHIFT transition following a RETAIN transition. This is due to the fact that a RETAIN transition ideally predicts a SMOOTH-SHIFT in the following utterance. In this case the SMOOTH-SHIFT is the "least effort" transition, because only the first element of the $C_f$ of the preceding utterance has to be checked to perform the SMOOTH-SHIFT transition, while in the case of CONTINUE at least one more check has to be performed. Hence, we claim that no one particular centering transition is preferred over another. Instead, we postulate that some *centering transition pairs* are preferred over others. Following this line of argumentation, we here propose to classify all occurrences of centering transition pairs with respect to the costs they imply. The cost-based evaluation of different $C_f$ orderings refers to evaluation criteria which form an intrinsic part of the centering model[6].

|  | Transition Types | naive | naive & ante > express | canonical | canonical & ante > express | functional |
|---|---|---|---|---|---|---|
| IT | CONTINUE | 49 | 167 | 102 | 197 | 309 |
|  | RETAIN | 269 | 158 | 226 | 131 | 25 |
|  | SMOOTH-SHIFT | 32 | 41 | 24 | 35 | 51 |
|  | ROUGH-SHIFT | 39 | 23 | 37 | 26 | 4 |
|  | Errors | 69 | 70 | 68 | 69 | 67 |
| Spiegel | CONTINUE | 17 | 28 | 37 | 43 | 50 |
|  | RETAIN | 42 | 32 | 28 | 23 | 12 |
|  | SMOOTH-SHIFT | 9 | 9 | 7 | 8 | 13 |
|  | ROUGH-SHIFT | 7 | 6 | 3 | 1 | 0 |
|  | Errors | 18 | 19 | 16 | 17 | 16 |
| Müller | CONTINUE | 31 | 31 | 32 | 32 | 36 |
|  | RETAIN | 19 | 19 | 18 | 18 | 15 |
|  | SMOOTH-SHIFT | 15 | 17 | 15 | 16 | 18 |
|  | ROUGH-SHIFT | 14 | 12 | 14 | 13 | 10 |
|  | Errors | 22 | 22 | 22 | 22 | 22 |
| Σ | CONTINUE | 97 | 226 | 171 | 272 | 395 |
|  | RETAIN | 330 | 209 | 272 | 172 | 52 |
|  | SMOOTH-SHIFT | 56 | 67 | 46 | 59 | 82 |
|  | ROUGH-SHIFT | 60 | 41 | 54 | 40 | 14 |
|  | Errors (specific errors) | 109 (10) | 111 (12) | 106 (7) | 108 (9) | 105 (6) |

Table 8: Numbers of Centering Transitions

Transition pairs hold for two immediately successive utterances. We distinguish between two types of transition pairs, *cheap* ones and *expensive* ones. We call a transition pair *cheap* if the *backward-looking center* of the current utterance is correctly predicted by the *preferred center* of the immediately preceding utterance, i.e., $C_b(U_i) = C_p(U_{i-1}), i = 2 \ldots n$. Transition pairs are called *expensive* if the *backward-looking center* of the current utterance is not correctly predicted by the *preferred center* of the immediately preceding utterance, i.e., $C_b(U_i) \neq C_p(U_{i-1}), i = 2 \ldots n$. Table 9 contains a detailed synopsis of cheap and expensive transition pairs. In particular, chains of the RETAIN transition in passages where the $C_b$ does not change (passages with constant theme) show that the canonical ordering constraints for the *forward-looking centers* are not appropriate.

The numbers of centering transition pairs generated by the different approaches are shown in Table 10. In general, the functional approach shows the best results, while the naive and the canonical approaches work reasonably well for the literary text, but exhibit a poor performance for the texts from the IT domain and the news magazine. The results for the latter approaches become only slightly more positive with the modification of ranking the antecedent of a textual ellipsis above the elliptical expression, but they do not compare to the results of the functional approach.

We were also interested in finding out whether the functional ordering we propose possibly "includes" the grammatical role based criteria discussed so far. We, therefore, re-evaluated the examples already annotated with $C_b/C_f$ data available in the literature (for the English language, we considered all examples from Grosz et al. (1995) and Brennan et al. (1987); for Japanese we took the data from Walker et al. (1994)). Surprisingly enough, all examples of Grosz et al. (1995) passed the test successfully. Only with respect to the troublesome *Alfa Romeo driving* scenario (cf. Brennan et al. (1987, p.157)) our constraints fail to properly rank the elements of the third sentence $C_f$ of that example.[7] Note also that these results were achieved without having recourse to extra constraints, e.g., the shared property constraint to account for anaphora parallelism (Kameyama, 1986).

We applied our constraints to Japanese examples in the same way. Again we abandoned all extra constraints set up in these studies, e.g., the Zero Topic Assignment (ZTA) rule and the special role of empathy verbs (Walker et al., 1994). However, the results our constraints generate are the same as those generated by Walker et al. including these model extensions. Only a single problematic case remains, *viz.* example (30) of Walker et al. (1994, p.214) causes the same problems they described (discourse-initial utterance, semantic or world knowledge should be available). Even for the crucial examples (32)-(36) of Walker et al. (1994, p.216-221) our constraints generate the same $C_f$s as Walker et al.'s constraints with ZTA.

---

[6]As a consequence of this postulate, we have to redefine Rule 2 of the Centering Constraints (Grosz et al., 1995, p.215) appropriately, which gives an informal characterization of a preference for sequences of CONTINUE over sequences of RETAIN and, similarly, sequences of RETAIN over sequences of SHIFT. Our specification for the case of text interpretation says that cheap transitions are preferred over expensive ones, with cheap and expensive transitions as defined in Table 9.

[7]In essence, the very specific problem addressed by that example seems to be that *Friedman* has not been previously introduced in the local discourse segment and is only accessible via the global focus.

|  | CONTINUE | RETAIN | SMOOTH-SHIFT | ROUGH-SHIFT |
|---|---|---|---|---|
| – | **cheap** | expensive | – | – |
| CONTINUE | **cheap** | **cheap** | expensive | expensive |
| RETAIN | expensive | expensive | **cheap** | expensive |
| SMOOTH-SHIFT | **cheap** | expensive | expensive | expensive |
| ROUGH-SHIFT | expensive | expensive | **cheap** | expensive |

Table 9: Costs for Transition Pairs

|  | cost type | naive | naive & ante > express | canonical | canonical & ante > express | functional |
|---|---|---|---|---|---|---|
| IT | cheap | 72 | 180 | 129 | 236 | 321 |
|  | expensive | 317 | 209 | 260 | 153 | 68 |
| Spiegel | cheap | 25 | 36 | 45 | 51 | 62 |
|  | expensive | 50 | 39 | 30 | 24 | 13 |
| Müller | cheap | 45 | 48 | 46 | 48 | 55 |
|  | expensive | 34 | 31 | 33 | 31 | 24 |
| $\Sigma$ | cheap | 142 | 264 | 220 | 335 | 438 |
|  | expensive | 401 | 279 | 323 | 208 | 105 |

Table 10: Cost Values for Centering Transition Pair Types

To summarize the results of our empirical evaluation, we first claim that our proposal based on functional criteria leads to substantially better and — with respect to the inference load placed on the text understander, whether human or machine — more plausible results for languages with free word order than the structural constraints given by Grosz et al. (1995) and those underlying a naive approach. We base these observations on an evaluation approach which considers transition pairs in terms of the inference load specific pairs imply. Second, we have gathered some evidence, still far from being conclusive, that the functional constraints on centering seem to incorporate the structural constraints for English and the modified structural constraints for Japanese. Hence, we hypothesize that functional constraints on centering might constitute a general mechanism for treating free <u>and</u> fixed word order languages by the same descriptive mechanism. This claim, however, has to be further substantiated by additional cross-linguistic empirical studies.

## 5 Comparison with Related Approaches

The centering model (Grosz et al., 1983; 1995) is concerned with the interactions between the local coherence of discourse and the choices of referring expressions. Crucial for the centering model is the way the forward-looking centers are organized. Despite several cross-linguistic studies a kind of "standard" has emerged based on the study of English (cf. Table 1 in Section 1). Only few of these cross-linguistic studies have led to changes in the basic order of discourse entities, the work of Walker et al. (1990; 1994) being the most far reaching exception. They consider the role of expressive means in Japanese to indicate topic status and the speaker's perspective, thus introducing *functional* notions, viz. TOPIC and EMPATHY, into the discussion. German, the object language we deal with, is also a free word order language like Japanese (possibly even more constrained). Our basic revision of the ordering scheme completely abandons grammatical role information and replaces it with entirely functional notions reflecting the information structure of the utterances in the discourse. Interestingly enough, several extra assumptions introduced to account, e.g., for anaphora parallelism (e.g., the shared property constraint formulated by Kameyama (1986)) can be eliminated without affecting the correctness of anaphora resolutions. Rambow (1993) has presented a theme/rheme distinction within the centering model to which we fully subscribe. His proposal concerning the centering analysis of German (already referred to as the "naive" approach; cf. Section 4) is limited, however, to the German middlefield and, hence, incomplete.

A common topic of criticism relating to focusing approaches to anaphora resolution has been the diversity of data structures they require, which are likely to hide the underlying linguistic regularities. Focusing algorithms prefer the discourse element already in focus for anaphora resolution, thus considering context-boundedness, too. But the items of the focus lists are either ordered by thematic roles (Sidner, 1983) or grammatical roles (Suri & McCoy, 1994; Dahl & Ball, 1990)). Dahl & Ball (1990) improve the focusing mechanism by simplifying its data structures and, thus, their proposal is more closely related to the centering model than any other focusing mechanism. But their approach still relies upon grammatical information for the ordering of the centering list, while we use only the functional information structure as the guiding principle.

# 6 Conclusion

In this paper, we provided an account for ordering the forward-looking centers which is entirely based on functional notions, grounded on the information structure of utterances in a discourse. We motivated our proposal by the constraints which hold for a free word order language such as German and derived our results from data-intensive empirical studies of (real-world) expository texts. We have gathered preliminary evidence that the functional ordering of discourse entities in the centers seems to coincide with the grammatical roles of fixed word order languages. We also augmented the ordering criteria of the forward-looking center such that it accounts not only for (pro)nominal but also for functional anaphora (textual ellipsis), an issue that, so far, has only been sketchily dealt with in the centering framework. The extensions we propose have been validated by the empirical analysis of real-world expository texts of considerable length. We thus follow methodological principles of corpus-based studies that have been successfully exercised in the work of Passonneau (1993). Still open are proper descriptions of deictic expressions, proper names (cf. the *Alfa Romeo driving* scenario), and plural or generic definite noun phrases. An anaphora resolution module and an ellipsis handler based on this functional centering model has been implemented as part of a comprehensive text parser for German.

**Acknowledgments.** We would like to thank our colleagues in the $\mathcal{CLIF}$ group for fruitful discussions and Jon Alcantara (Cambridge, UK) for re-reading the final version via Internet. This work has been funded by *LGFG Baden-Württemberg* (M. Strube).

## References

Brennan, S. E., M. W. Friedman & C. J. Pollard (1987). A centering approach to pronouns. In *Proc. of ACL-87*, pp. 155–162.

Clark, H. H. (1975). Bridging. In *Proc. of TINLAP-1*, pp. 169–174.

Cote, S. (1996). Ranking forward-looking centers. In E. Prince, A. Joshi & M. Walker (Eds.), *Centering in Discourse*. Oxford: Oxford University Press.

Dahl, D. A. & C. N. Ball (1990). Reference resolution in PUNDIT. In P. Saint-Dizier & S. Szpakowicz (Eds.), *Logic and Logic Grammars for Language Processing*, pp. 168–184. Chichester/UK: Ellis Horwood.

Dahl, Ö. (Ed.) (1974). *Topic and Comment, Contextual Boundness, and Focus.* Hamburg: Buske.

Daneš, F. (1974a). Functional sentence perspective and the organization of the text. In F. Daneš (Ed.), *Papers on Functional Sentence Perspective*, pp. 106–128. Prague: Academia.

Daneš, F. (Ed.) (1974b). *Papers on Functional Sentence Perspective.* Prague: Academia.

Gordon, P. C., B. J. Grosz & L. A. Gilliom (1993). Pronouns, names, and the centering of attention in discourse. *Cognitive Science*, 17:311–347.

Grosz, B. J., A. K. Joshi & S. Weinstein (1983). Providing a unified account of definite noun phrases in discourse. In *Proc. of ACL-83*, pp. 44–50.

Grosz, B. J., A. K. Joshi & S. Weinstein (1995). Centering: A framework for modeling the local coherence of discourse. *Computational Linguistics*, 21(2):203–225.

Hahn, U. & M. Strube (1996). Incremental centering and center ambiguity. In *Proc. of the $18^{th}$ Annual Conf. of the Cognitive Science Society; La Jolla, CA*.

Hahn, U., M. Strube & K. Markert (1996). Bridging textual ellipses. In *Proc. of COLING-96*.

Hajičová, E., V. Kuboň & P. Kuboň (1992). Stock of shared knowledge: A tool for solving pronominal anaphora. In *Proc. of COLING-92*, Vol. 1, pp. 127–133.

Hajičová, E., H. Skoumalová & P. Sgall (1995). An automatic procedure for topic-focus identification. *Computational Linguistics*, 21(1):81–94.

Halliday, M. A. K. (1967). Notes on transitivity and theme in English, Part 2. *Journal of Linguistics*, 3:199–244.

Halliday, M. A. K. & R. Hasan (1976). *Cohesion in English.* London: Longman.

Kameyama, M. (1986). A property-sharing constraint in centering. In *Proc. of ACL-86*, pp. 200–206.

Passonneau, R. J. (1993). Getting and keeping the center of attention. In M. Bates & R. Weischedel (Eds.), *Challenges in Natural Language Processing*, pp. 179–227. Cambridge, UK: Cambridge University Press.

Rambow, O. (1993). Pragmatic aspects of scrambling and topicalization in German. In *IRCS Workshop on Centering in Discourse. Univ. of Pennsylvania, 1993*.

Sidner, C. L. (1983). Focusing in the comprehension of definite anaphora. In M. Brady & R. Berwick (Eds.), *Computational Models of Discourse*, pp. 267–330. Cambridge, MA: MIT Press.

Strube, M. (1996). Processing complex sentences in the centering framework. In *Proc. of ACL-96*.

Strube, M. & U. Hahn (1995). *ParseTalk* about sentence- and text-level anaphora. In *Proc. of EACL-95*, pp. 237–244.

Suri, L. Z. & K. F. McCoy (1994). RAFT/RAPR and centering: A comparison and discussion of problems related to processing complex sentences. *Computational Linguistics*, 20(2):301–317.

Turan, U. (1995). *Null vs. Overt Subjects in Turkish: A Centering Approach.*, (Ph.D. thesis). University of Pennsylvania.

Walker, M. A., M. Iida & S. Cote (1990). Centering in Japanese discourse. In *Proc. of COLING-90, Appendix, 6pp*.

Walker, M. A., M. Iida & S. Cote (1994). Japanese discourse and the process of centering. *Computational Linguistics*, 20(2):193–233.